\pdfoutput=1

\documentclass[11pt]{article}

\usepackage[]{acl}
\usepackage{times}
\usepackage{latexsym}
\usepackage{enumitem}

\usepackage[T1]{fontenc}
\usepackage{siunitx}

\usepackage[utf8]{inputenc}

\usepackage{longtable}
\usepackage{supertabular}
\usepackage{amsmath}
\usepackage{lipsum}

\usepackage{enumitem}
\usepackage{amsmath}
\usepackage{color}

\usepackage{soul}

\title{HLDC: Hindi Legal Documents Corpus}

\usepackage{graphicx}
\usepackage{array}
\usepackage{multirow}

\usepackage{float}

\usepackage{listings}
\usepackage{devanagari}
\usepackage{booktabs}

 \author{Arnav Kapoor$^\dagger$, Mudit Dhawan$^\ddagger$, Anmol Goel$^\dagger$, \\ {\bf T.H. Arjun}$^\dagger$,  {\bf Akshala Bhatnagar}$^\ddagger$,  
 {\bf Vibhu Agrawal}$^\ddagger$, {\bf Amul Agrawal}$^\dagger$, \\
 {\bf Arnab Bhattacharya}$^\mathparagraph$,   
 {\bf Ponnurangam Kumaraguru}$^\dagger$, 
 {\bf Ashutosh Modi}$^\mathparagraph$\thanks{\ \ Corresponding Author} \\ 
        $^\dagger$IIIT Hyderabad,
        $^\ddagger$IIIT Delhi,
        $^\mathparagraph$IIT Kanpur\\
  \texttt{\small \{arnav.kapoor,anmol.goel,arjun.thekoot\}@research.iiit.ac.in} \\
  \texttt{\small \{mudit18159,akshala18012,vibhu18116\}@iiitd.ac.in}, \texttt{\small amul.agrawal@students.iiit.ac.in},\\
  \texttt{\small arnabb@cse.iitk.ac.in, pk.guru@iiit.ac.in, ashutoshm@cse.iitk.ac.in}\\}

\begin{document}
\maketitle
\begin{abstract}
Many populous countries including India are burdened with a considerable backlog of legal cases. Development of automated systems that could process legal documents and augment legal practitioners can mitigate this. However, there is a dearth of high-quality corpora that is needed to develop such data-driven systems. The problem gets even more pronounced in the case of low resource languages such as Hindi. In this resource paper, we introduce the \emph{Hindi Legal Documents Corpus (HLDC)}, a corpus of more than 900K legal documents in Hindi. Documents are cleaned and structured to enable the development of downstream applications. Further, as a use-case for the corpus, we introduce the task of bail prediction. We experiment with a battery of models and propose a Multi-Task Learning (MTL) based model for the same. MTL models use summarization as an auxiliary task along with bail prediction as the main task. Experiments with different models are indicative of the need for further research in this area. We release the corpus and model implementation code with this paper: {\url{https://github.com/Exploration-Lab/HLDC}}. 
\end{abstract}
\section{Introduction}
\label{sec:introduction}

In recent times, the legal system in many populous countries (e.g., India) has been inundated with a large number of legal documents and pending cases \cite{backlog-cases2019}. There is an imminent need for automated systems to process legal documents and help augment the legal procedures. For example, if a system could readily extract the required information from a legal document for a legal practitioner, then it would help expedite the legal process. However, the processing of legal documents is challenging and is quite different from conventional text processing tasks. For example, legal documents are typically quite long (tens of pages), highly unstructured and noisy (spelling and grammar mistakes since these are typed), use domain-specific language and jargon; consequently, pre-trained language models do not perform well on these \cite{malik-etal-2021-ildc}. Thus, to develop legal text processing systems and address the challenges associated with the legal domain, there is a need for creating specialized legal domain corpora.

In recent times, there have been efforts to develop such corpora. For example, \citet{chalkidis-etal-2019-neural} have developed an English corpus of European Court of Justice documents, while \citet{malik-etal-2021-ildc} have developed an English corpus of Indian Supreme Court documents.  \citet{xiao2018cail2018} have developed Chinese Legal Document corpus. However, to the best of our knowledge, there does not exist any legal document corpus for the Hindi language (a language belonging to the Indo-European family and pre-dominantly spoken in India). Hindi uses Devanagari script \cite{enwiki:1053977349} for the writing system. Hindi is spoken by approximately 567 million people in the world \cite{hindi-worlddata}. Most of the lower (district) courts in northern India use Hindi as the official language. However, most of the legal NLP systems that currently exist in India have been developed on English, and these do not work on Hindi legal documents \cite{malik-etal-2021-ildc}. To address this problem, in this paper, we release a large corpus of Hindi legal documents (\textsc{Hindi Legal Documents Corpus} or \textsc{HLDC}) that can be used for developing NLP systems that could augment the legal practitioners by automating some of the legal processes. Further, we show a use case for the proposed corpus via a new task of \emph{bail prediction}. 

India follows a Common Law system and has a three-tiered court system with District Courts (along with Subordinate Courts) at the lowest level (districts), followed by High Courts at the state level, and the Supreme Court of India at the highest level. In terms of number of cases, district courts handle the majority. According to India's National Judicial Data Grid, as of November 2021, there are approximately 40 million cases pending in District Courts \cite{njdc-district} as opposed to 5 million cases pending in High Courts. These statistics show an immediate need for developing models that could address the problems at the grass-root levels of the Indian legal system. Out of the 40 million pending cases, approximately 20 million are from courts where the official language is Hindi \cite{njdc-district}. In this resource paper, we create a large corpus of 912,568 Hindi legal documents. In particular, we collect documents from the state of Uttar Pradesh, the most populous state of India with a population of approximately 237 million \cite{up-population}. The Hindi Legal Documents Corpus (HLDC) can be used for a number of legal applications, and as a use case, in this paper, we propose the task of \textit{Bail Prediction}.

Given a legal document with facts of the case, the task of \emph{bail prediction} requires an automated system to predict if the accused should be granted bail or not. The motivation behind the task is not to replace a human judge but rather augment them in the judicial process. Given the volume of cases, if a system could present an initial analysis of the case, it would expedite the process. As told to us by legal experts and practitioners, given the economies of scale, even a small improvement in efficiency would result in a large impact. We develop baseline models for addressing the task of bail prediction.

In a nutshell, we make the following main contributions in this resource paper:
\begin{itemize}[noitemsep,topsep=0pt,leftmargin=*]
\item We create a \emph{Hindi Legal Documents Corpus (HLDC)} of 912,568  documents. These documents are cleaned and structured to make them usable for downstream NLP/IR applications. Moreover, this is a growing corpus as we continue to add more legal documents to HLDC. We release the corpus and model implementation code with this paper: {\url{https://github.com/Exploration-Lab/HLDC}}.  
\item As a use-case for applicability of the corpus for developing legal systems, we propose the task of \emph{Bail Prediction}.
\item For the task of bail prediction, we experiment with a variety of deep learning models. We propose a multi-task learning model based on transformer architecture. The proposed model uses extractive summarization as an auxiliary task and bail prediction as the main task.   
\end{itemize}

\section{Related Work}
\label{sec:related-work}

In recent years there has been active interest in the application of NLP techniques to the legal domain \cite{zhong-etal-2020-nlp}. A number of tasks and models have been proposed, inter alia, Legal Judgment Prediction \cite{chalkidis-etal-2019-neural}, Legal Summarization \cite{azzopardi-comparative-2019,vu2019}, Prior Case Retrieval \cite{JACKSON2003239,shao-bert-pli-2020}, Legal Question Answering \cite{kim-two-step-2017}, Catchphrase Extraction \cite{catchphrase-galgani},  Semantic Segmentation \cite{rr-kalamakar-2022,rr-malik-2022}.

Legal Judgement Prediction (LJP) involves predicting the final decision from the facts and arguments of the case. \citet{chalkidis-etal-2019-neural} released 11,478 cases from the European Court of Human Rights (ECHR). It contains facts, articles violated (if any), and the importance scores. \citet{malik-etal-2021-ildc} provided 34,816 case documents from the Supreme Court of India for the prediction task. \citet{strickson-legal-2020} published 4,959 documents from the U.K.’s Supreme court (the highest court of appeal). 

Majority of corpora for Legal-NLP tasks have been in English; recently, there have been efforts to address other languages as well, for example, \citet{xiao2018cail2018}, have created a large-scale Chinese criminal judgment prediction dataset with over 2.68 million legal documents. Work on Legal-NLP in languages other than English is still in its incipient stages. Our paper contributes towards these efforts by releasing corpus in Hindi. 

Majority of the work in the legal domain has focused on the higher court \cite{malik-etal-2021-ildc,strickson-legal-2020,zhong2019jec}; however, the lower courts handle the maximum number of cases. We try to address this gap by releasing a large corpus of district court level legal documents. 

\begin{figure*}[t]
\includegraphics[width=\textwidth]{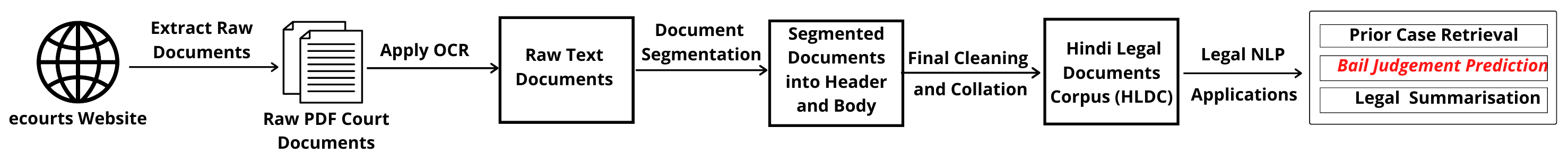}
\centering
\caption{HLDC corpus creation pipeline}
\label{fig:data-pipeline}
\end{figure*}

Some of the recent work has explored other Legal-NLP tasks in languages other than English. \citet{chalkidis-etal-2021-multieurlex} released a multilingual dataset of 65K European Union (E.U.) laws for topic classification of legal documents. The data was translated into the 23 official E.U. languages and annotated with labels from the multilingual thesaurus, EUROVOC. 
\citet{luz-etal-propor2018} have released 70 documents in Portuguese for Legal Named Entity Recognition. The dataset contains specific tags for law and legal cases entities in addition to the normal tags for named entities like person,  locations, organisation and time-entities. 
COLIEE (Competition on Legal Information Extraction/Entailment) tasks  \cite{kojima-coliee-2018,Kano2017OverviewOC} have published legal data in Japanese, along with their English translation. The competition has two sub-tasks, a legal information retrieval task and an entailment identification task between law articles and queries.    
Multiple datasets in Chinese have been released for different tasks, namely Reading Comprehension \cite{sun-cjrc:2019}, Similar Case Matching \cite{Xiao2019CAIL2019SCMAD}, Question Answering \cite{zhong2019jec}.
\citet{sun-cjrc:2019} proposed Chinese judicial reading comprehension (CJRC) dataset with about 10K documents and almost 50K questions with answers. \citet{zhong2019jec} presented JEC-QA, a legal question answering dataset collected from the National Judicial Examination of China with about 26K multiple-choice questions. They augment the dataset with a database
containing the legal knowledge required 
to answer the questions and also assign meta information to each of the questions for in-depth analysis. \citet{Xiao2019CAIL2019SCMAD} proposed CAIL2019-SCM, a dataset containing 8,964 triplets of the case document, with the objective to identify which two cases are more similar in the triplets. Similar case matching has a crucial application as it helps to identify comparable historical cases. A historical case with similar facts often serves as a legal precedent and influences the judgement. Such historical information can be used to make the legal judgement prediction models more robust.   
 
\citet{kleinberg-human-2017} proposed bail decision prediction as a good proxy to gauge if machine learning can improve human decision making. A large number of bail documents along with the binary decision (granted or denied) makes it an ideal task for automation. In this paper, we also propose the bail prediction task using the HLDC corpus.

\section{Hindi Legal Documents Corpus}
\label{sec:hldc}

\begin{figure}[t]
\includegraphics[scale=0.50]{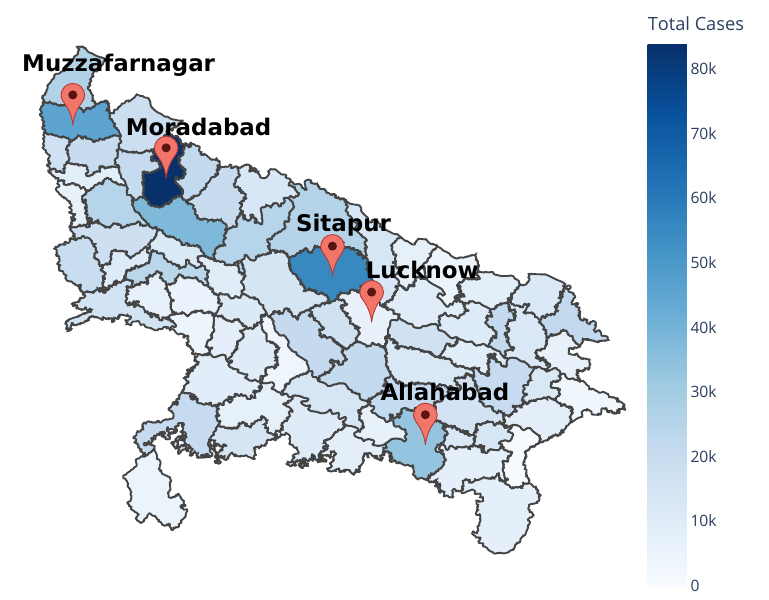}
\centering
\caption{Variation in number of case documents per district in the state of U.P. Prominent districts are marked.}
\label{fig:case-numbers-in-up}
\end{figure}

Hindi Legal Documents Corpus (HLDC) is a corpus of 912,568 Indian legal case documents in the Hindi language. The corpus is created by downloading data from the e-Courts website (a publicly available website: \url{https://districts.ecourts.gov.in/}). All the legal documents we consider are in the public domain. We download case documents pertaining to the district courts located in the Indian northern state of Uttar Pradesh (U.P.). We focus mainly on the state of U.P. as it is the most populous state of India, resulting in the filing of a large number of cases in district courts. U.P. has 71 districts and about 161 district courts. U.P. is a predominantly Hindi speaking state, and consequently, the official language used in district courts is Hindi. We crawled case documents from all districts of U.P. corresponding to cases filed over two years, from May 01, 2019 to May 01, 2021. Figure \ref{fig:case-numbers-in-up} shows the map of U.P. and district wise variation in the number of cases. As can be seen in the plot, the western side of the state has more cases; this is possibly due to the high population and more urbanization in the western part. Table \ref{tab:case-types-in-up} shows \%-wise division of different case types in HLDC. As evident from the table, majority of documents pertain to bail applications. HLDC corpus has a total of 3,797,817 unique tokens, and on average, each document has 764 tokens. 

\noindent\textbf{HLDC Creation Pipeline:} We outline the entire pipeline used to create the corpus in Figure \ref{fig:data-pipeline}. The documents on the website are originally typed in Hindi (in Devanagari script) and then scanned to PDF format and uploaded. The first step in HLDC creation is the downloading of documents from the e-Courts website. We downloaded a total of 1,221,950 documents. To extract Hindi text from these, we perform OCR (Optical Character Recognition) via the Tesseract tool\footnote{\url{https://github.com/tesseract-ocr}}.  Tesseract worked well for our use case as the majority of case documents were well-typed, and it outperformed other OCR libraries\footnote{\url{https://github.com/JaidedAI/EasyOCR}}. The obtained text documents were further cleaned to remove noisy documents, e.g. too short (< 32 bytes) or too long (> 8096 bytes) documents, duplicates, and English documents (details in Appendix \ref{app:data-cleaning}). This resulted in a total of 912,568 documents in HLDC. We anonymized the corpus with respect to names and locations. We used a gazetteer\footnote{\url{https://github.com/piyusharma95/NER-for-Hindi}, \url{https://github.com/balasahebgulave/Dataset-Indian-Names}} along with regex-based rules for NER to anonymize the data. List of first names, last names, middle names, locations, titles like {\dn p\2EXt} (\texttt{Pandit}: title of Priest), {\dn srjF} (\texttt{Sir}: Sir), month names and day names were normalized to \textless{\dn nAm}\textgreater{} (\texttt{Naam}: <name>). The gazetteer also had some common ambiguous words (these words can be names or sometimes verbs) like {\dn \3FEwAT\0nA} (\texttt{Prathna}: Can refer to prayer, the action of request or name), {\dn gyA} (\texttt{Gaya}: can refer to location name or verb), {\dn EkyA} (\texttt{Kiya}: can refer to infinitive `to do' or name), {\dn ElyA} (\texttt{Liya}: can refer to infinitive `to take' or name). These were removed. 
Further, we ran RNN-based Hindi NER model\footnote{\url{https://github.com/flairNLP/flair}} on a subset of documents to find additional entities and these were subsequently used to augment our gazetteer (details Appendix \ref{ner-removal}). Phone numbers were detected using regex patterns and replaced with a \textless{\dn \327won{\rs -\re}n\2br}\textgreater{} (<phone-number>) tag, numbers written in both English and Hindi were considered. 

Legal documents, particularly in lower courts, are highly unstructured and lack standardization with respect to format and sometimes even the terms used. We converted the unstructured documents to semi-structured documents. We segmented each document into a \textit{header} and a \textit{body}. The header contains the meta-information related to the case, for example, case number, court identifier, and applicable sections of the law. The body contains the facts of the case, arguments, judge's summary, case decision and other information related to the final decision. The documents were segmented using regex and rule based approaches as described in Appendix \ref{app:bail-document-segmentation}.

\textbf{Case Type Identification:} HLDC documents were processed to obtain different case types (e.g., Bail applications, Criminal Cases). The case type was identified via the meta-data that comes with each document. However, different districts use a variation of the same case type name (e.g., Bail Application vs Bail App.). We resolved these standardization issues via manual inspection and regex-based patterns, resulting in a final list of 300 unique case types.

\begin{table}[t]
\small
    \centering
    \begin{tabular}{p{0.65\columnwidth}r}
        \toprule
        \bf Case Type & \bf \% in HLDC\\
        \midrule
        Bail Applications & 31.71 \\ 
        Criminal Cases & 20.41 \\ 
        Original Suits & 6.54 \\ 
        Warrant or Summons in Criminal Cases & 5.24 \\ 
        Complaint Cases & 4.37 \\ 
        Civil Misc & 3.4 \\ 
        Final Report & 3.32 \\ 
        Civil Cases & 3.23 \\ 
        Others (Matrimonial Cases, Session Trial, Motor Vehicle Act, etc.) & 21.75 \\ 
        \bottomrule
    \end{tabular}
    \caption{Case types in HLDC. Out of around 300 different case types, we only show the prominent ones. Majority of the case documents correspond to bail applications.}
    \label{tab:case-types-in-up}
\end{table}

\begin{figure}[t]
\includegraphics[scale=0.5]{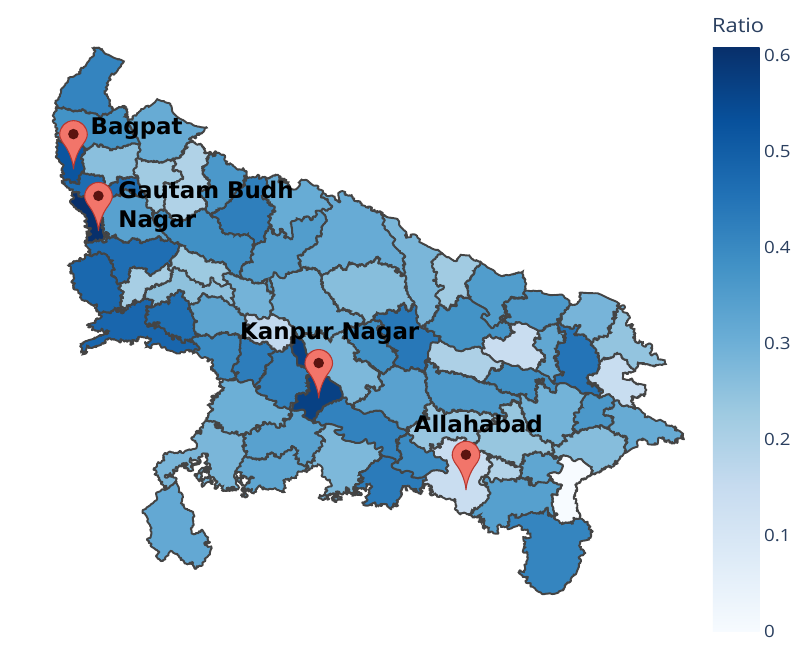}
\centering
\caption{Ratio of number of bail applications to total number of applications in U.P.} 
\label{fig:bail-ratios-in-up}
\end{figure}

\begin{figure*}[t]
\includegraphics[scale=0.30]{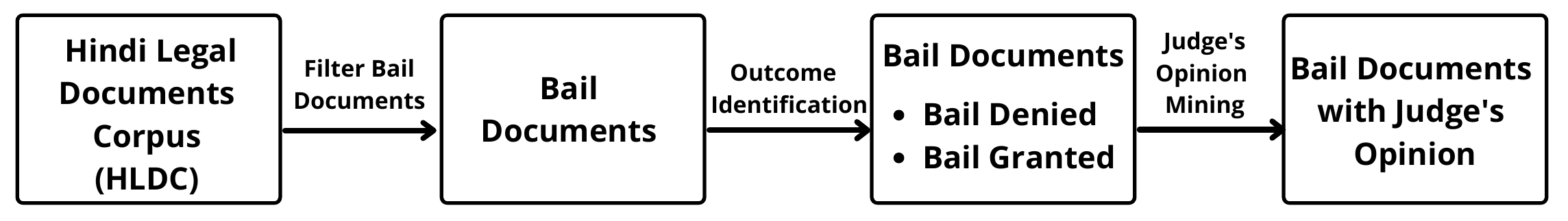}
\centering
\caption{Bail Corpus Creation Pipeline}
\label{fig:bail_pipeline}
\end{figure*}


\textbf{Lexical Analysis:} Although Hindi is the official language, U.P. being a large and populous state, has different dialects of Hindi spoken across the state.  We found evidence of this even in official legal documents.  For example, the word {\dn sAEkn} (\texttt{Sakin}: motionless) appears 11,614 times in the dataset, 63.8\% occurrences of the word come from 6 districts of East U.P. (Ballia, Azamgarh, Maharajganj, Deoria, Siddharthnagar and Kushinagar). 
This particular variant of motionless being used most often only in East U.P. 
Similarly, the word {\dn gOv\2fFy} (\texttt{Gaushiya}: cow and related animals) is mostly used in North-Western UP (Rampur, Pilibhit, Jyotiba Phule Nagar (Amroha), Bijnor, Budaun, Bareilly, Moradabad). 
Three districts - Muzaffarnagar, Kanshiramnagar and Pratapgarh district constitute 81.5\% occurrences of the word {\dn d\2X} (\texttt{Dand}: punishment). These districts are, however, spread across UP. An important thing to note is that words corresponding to specific districts/areas are colloquial and not part of the standard Hindi lexicon. This makes it difficult for prediction model to generalize across districts (\S \ref{sec:results}).  

\noindent\textbf{Corpus of Bail Documents:} 
Bail is the provisional release of a suspect in any criminal offence on payment of a bail bond and/or additional restrictions. Bail cases form a large majority of cases in the lower courts, as seen in Table \ref{tab:case-types-in-up}. Additionally, they are very time-sensitive as they require quick decisions. 
For HLDC, the ratio of bail documents to total cases in each district is shown in Figure \ref{fig:bail-ratios-in-up}. As a use-case for the corpus, we further investigated the subset of the corpus having only the bail application documents (henceforth, we call it Bail Corpus). 

\textbf{Bail Document Segmentation:} For the bail documents, besides the header and body, we further segmented the body part into more sub-sections (Figure \ref{fig:bail_pipeline}). Body is further segmented into \textbf{Facts and Arguments}, \textbf{Judge's summary} and \textbf{Case Result}. Facts contain the facts of the case and the defendant and prosecutor's arguments. Most of the bail documents have a concluding paragraph where the judge summarizes their viewpoints of the case, and this constitutes the judge's summary sub-section. The case result sub-section contains the final decision given by the judge. More details about document segmentation are in Appendix \ref{app:bail-document-segmentation}.



\textbf{Bail Decision Extraction}: Decision was extracted from Case Result Section using a rule based approach (Details in Appendix \ref{app:bail-decision}).

\textbf{Bail Amount Extraction}: If bail was granted, it usually has some bail amount associated with it. We extracted this bail amount 
using regex patterns (Details in Appendix \ref{app:bail-amount}). 

We verified each step of the corpus creation pipeline (Detailed analysis in Appendix \ref{app:pipeline}) to ensure the quality of the data. We initially started with 363,003 bail documents across all the 71 districts of U.P., and after removing documents having segmentation errors, we have a Bail corpus with 176,849 bail documents. The bail corpus has a total of 2,342,073 unique tokens, and on average, each document has 614 tokens. A sample document segmented into various sections is shown in Appendix \ref{app:sample-example}.

\section{HLDC: Ethical Aspects}

We create HLDC to promote research and automation in the legal domain dealing with under-researched and low-resource languages like Hindi. The documents that are part of HLDC are in the public domain and hence accessible to all. Given the volume of pending cases in the lower courts, our efforts are aimed towards improving the legal system, which in turn would be beneficial for millions of people. Our work is in line with some of the previous work on legal NLP, e.g., legal corpora creation and legal judgement prediction (section \ref{sec:related-work}). Nevertheless, we are aware that if not handled correctly, legal AI systems developed on legal corpora can negatively impact an individual and society at large.
Consequently, we took all possible steps to remove any personal information and biases in the corpus. We anonymized the corpus (section \ref{sec:hldc}) with respect to names, gender information, titles, locations, times, judge's name, petitioners and appellant's name. As observed in previous work \cite{malik-etal-2021-ildc}, anonymization of a judge's name is important as there is a correlation between a case outcome and a judge name. Along with the HLDC, we also introduce the task of Bail Prediction. Bail applications constitute the bulk of the cases (\S\ref{sec:hldc}), augmentation by an AI system can help in this case. The bail prediction task aims not to promote the development of systems that replace humans but rather the development of systems that augment humans. The bail prediction task provides only the facts of the case to predict the final decision and avoids any biases that may affect the final decision. Moreover, the Bail corpus and corresponding bail prediction systems can promote the development of explainable systems \cite{malik-etal-2021-ildc}, we leave research on such explainable systems for future work. The legal domain is a relatively new area in NLP research, and more research and investigations are required in this area, especially concerning biases and societal impacts; for this to happen, there is a need for corpora, and in this paper, we make initial steps towards these goals. 

\section{Bail Prediction Task} \label{sec:bail-prediction}

To demonstrate a possible applicability for HLDC, we propose the \textit{Bail Prediction Task}, where given the facts of the case, the goal is to predict whether the bail would be granted or denied. Formally, consider a corpus of bail documents $\mathcal{D} = {b_1, b_2, \cdots, b_i}$, where each bail document is segmented as $b_i = (h_i, f_i, j_i, y_i)$. Here, $h_i, f_i, j_i$ and $y_i$ represent the header, facts, judge's summary and bail decision of the document respectively. Additionally, the facts of every document contain $k$ sentences, more formally, $f_i = (s_{i}^{1}, s_{i}^{2}, \cdots, s_{i}^{k})$, where $s_{i}^{k}$ represents the $k^{th}$ sentence of the $i^{th}$ bail document. We formulate the bail prediction task as a binary  classification problem. We are interested in modelling $p_{\theta}(y_i|f_i)$, which is the probability of the outcome $y_i$ given the facts of a case $f_i$. Here, $y_i \in \{0, 1\}$, i.e., 0 if bail is denied or 1 if bail is granted.

\section{Bail Prediction Models}

\begin{figure*}[t]
    \centering
    \includegraphics[width=\textwidth]{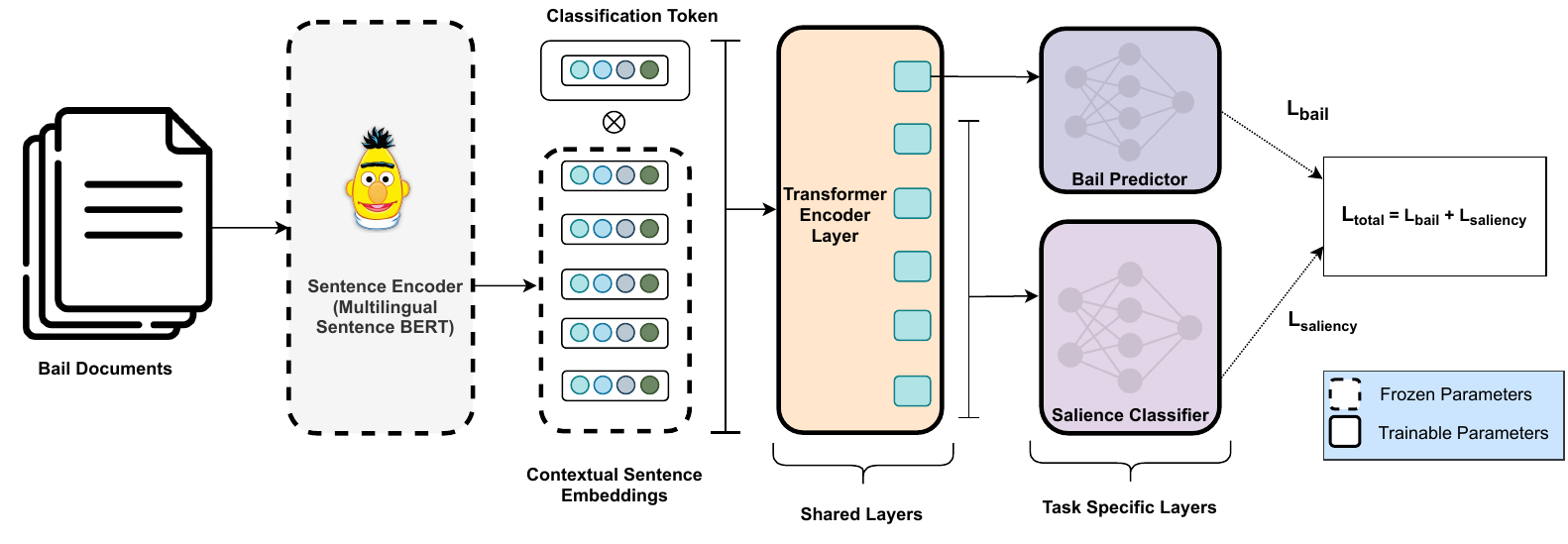}
    \caption{Overview of our multi-task learning approach.}
    \label{fig:MTL_model}
\end{figure*}

We initially experimented with off-the-shelf pre-trained models trained on general-purpose texts. However, as outlined earlier (\S\ref{sec:introduction}), the legal domain comes with its own challenges, viz. specialized legal lexicon, long documents, unstructured and noisy texts. Moreover, our corpus is from an under-resourced language (Hindi). Nevertheless, we experimented with existing fine-tuned (pre-trained) models and finally propose a multi-task model for the bail prediction task.  


\subsection{Embedding Based Models}
We experimented with classical embedding based model Doc2Vec \cite{doc2vec} and transformer-based contextualized embeddings model IndicBERT \cite{kakwani-etal-2020-indicnlpsuite}. Doc2Vec embeddings, in our case, is trained on the train set of our corpus. The embeddings go as input to SVM and XgBoost classifiers. IndicBERT is a transformer language model trained on 12 major Indian languages. However, IndicBERT, akin to other transformer LMs, has a limitation on the input's length (number of tokens). Inspired by \citet{malik-etal-2021-ildc, chalkidis-etal-2019-neural}, we experimented with fine-tuning IndicBERT in two settings: the first 512 tokens and the last 512 tokens of the document. The fine-tuned transformer with a classification head is used for bail prediction. 
    
\subsection{Summarization Based Models}
Given the long lengths of the documents, we experimented with prediction models that use summarization as an intermediate step. In particular, an extractive summary of a document goes as input to a fine-tuned transformer-based classifier (IndicBERT). Besides reducing the length of the document, extractive summarization helps to evaluate the salient sentences in a legal document and is a step towards developing explainable models. We experimented with both unsupervised and supervised extractive summarization models. 

For unsupervised approaches we experimented with TF-IDF \cite{tfidf-summary} and TextRank (a graph based method for extracting most important sentences) \cite{mihalcea2004textrank}. For the supervised approach, inspired by \citet{bajaj-etal-2021-long}, we propose the use of sentence salience classifier to extract important sentences from the document. Each document ($b_{i} = (h_i, f_i, j_i, y_i)$, \S\ref{sec:bail-prediction}) comes with a judge's summary $j_{i}$. For each sentence in the facts of the document ($f_{i}$) we calculate it's cosine similarity with judge's summary ($j_{i}$). Formally, salience of $k^{th}$ sentence $s_{i}^{k}$ is given by: $salience(s_{i}^{k}) = cos(h_{j_{i}}, h_{s_{i}^{k}})$. Here $h_{j_{i}}$ is contextualized distributed representation for $j_{i}$ obtained using multilingual sentence encoder \cite{sbert-multilingual}. Similarly, $h_{s_{i}^{k}}$ is the representation for the sentence $s_{i}^{k}$. The cosine similarities provides ranked list of sentences and we select top 50\% sentences as salient. The salient sentences are used to train (and fine-tune) IndicBERT based classifier. 

\subsection{Multi-Task Learning (MTL) Model}
As observed during experiments, summarization based models show improvement in results (\S\ref{sec:results}). Inspired by this, we propose a multi-task framework (Figure \ref{fig:MTL_model}), where bail prediction is the main task, and sentence salience classification is the auxiliary task. The intuition is that predicting the important sentences via the auxiliary task would force the model to perform better predictions and vice-versa. Input to the model are sentences corresponding to the facts of a case: $s_{i}^{1}, s_{i}^{2}, \ldots, s_{i}^{k}$. A multilingual sentence encoder \cite{sbert-multilingual} is used to get contextualized representation of each sentence: $\{ h_i^1, h_i^2, \cdots, h_i^k \}$. In addition, we append the sentence representations with a special randomly initialized CLS embedding \cite{devlin-etal-2019-bert} that gets updated during model training. The CLS and sentence embeddings are fed into standard single layer transformer architecture (shared transformer).  

\subsubsection{Bail Prediction Task}
A classification head (fully connected layer MLP) on the top of transformer CLS embedding is used to perform bail prediction. We use standard cross-entropy loss ($L_{bail}$) for training.  

\subsubsection{Salience Classification Task}
We use the salience prediction head (MLP) on top of sentence representations at the output of the shared transformer. For training the auxiliary task, we use sentence salience scores obtained via cosine similarity (these come from supervised summarization based model). For each sentence, we use binary-cross entropy loss ($L_{salience}$) to predict the salience. 

Based on our empirical investigations, both the losses are equally weighted, and total loss is given by $L = L_{bail} + L_{salience}$

\section{Experiments and Results} \label{sec:results}



\begin{table}[t]
\small
\centering
\resizebox{\columnwidth}{!}{
\begin{tabular}{|l|l|l|l|l|}
\hline
 &  & \textbf{Granted} & \textbf{Dismissed} & \textbf{Total} \\ \hline
\multirow{3}{*}{\begin{tabular}[c]{@{}l@{}}All \\ Districts\end{tabular}} & Train & 77010 (62\%) & 46732 (38\%) & 123742 \\ \cline{2-5}
 & Test & 21977 (62\%) & 13423 (38\%) & 35400 \\ \cline{2-5}
 & Validation & 11067 (63\%) & 6640 (37\%) & 17707 \\ \hline
\multirow{3}{*}{\begin{tabular}[c]{@{}l@{}}District\\ Wise\end{tabular}} & \begin{tabular}[c]{@{}l@{}}Train\\ (44 districts)\end{tabular} & 77220 (62\%) & 47121 (38\%) & 124341 \\ 
\cline{2-5}
 & \begin{tabular}[c]{@{}l@{}}Validation\\ (10 districts)\end{tabular} & 9563 (60\%) & 6366 (40\%) & 15929 \\ 
\cline{2-5}
 & \begin{tabular}[c]{@{}l@{}}Test\\ (17 districts)\end{tabular} & 23271 (64\%) & 13308 (36\%) & 36579 \\ 
\hline
\end{tabular}
}
\caption{Number of documents across each split}
\label{tab:data_splits}
\end{table}


\subsection{Dataset Splits}
We evaluate the models in two settings: all-district performance and district-wise performance. For the first setting, the model is trained and tested on the documents coming from all districts. The train, validation and test split is 70:10:20.  The district-wise setting is to test the generalization capabilities of the model. In this setting, the documents from 44 districts (randomly chosen) are used for training. Testing is done on a different set of 17 districts not present in train set. The validation set has another set of 10 districts. This split corresponds to a 70:10:20 ratio. Table \ref{tab:data_splits} provides the number of documents across splits. The corpus is unbalanced for the prediction class with about 60:40 ratio for positive to negative class (Table \ref{tab:data_splits}). All models are evaluated using standard accuracy and F1-score metric (Appendix \ref{eval-metrics}).




\begin{table}[t]
\small
\begin{tabular}{|l|ll|ll|}
\hline  
\multirow{2}{*}{Model Name}                      & \multicolumn{2}{l|}{District-wise} & \multicolumn{2}{l|}{All Districts} \\ \cline{2-5} 
                                                 & \multicolumn{1}{l|}{Acc.}  & F1      & \multicolumn{1}{l|}{Acc.}  & F1     \\ \hline
Doc2Vec + SVM                                    & \multicolumn{1}{l|}{0.72}    & 0.69  & \multicolumn{1}{l|}{0.79}    & 0.77 \\ \hline
Doc2Vec + XGBoost                                & \multicolumn{1}{l|}{0.68}    & 0.59  & \multicolumn{1}{l|}{0.67}    & 0.57 \\ \hline
IndicBert-First 512                           & \multicolumn{1}{l|}{0.65}    & 0.62  & \multicolumn{1}{l|}{0.73}    & 0.71 \\ \hline
IndicBert-Last 512                            & \multicolumn{1}{l|}{0.62}    & 0.60  & \multicolumn{1}{l|}{0.78}    & 0.76 \\ \hline
TF-IDF+IndicBert                           & \multicolumn{1}{l|}{0.76}    & 0.74  & \multicolumn{1}{l|}{\textbf{0.82}}    & \textbf{0.81} \\ \hline
TextRank+IndicBert                       & \multicolumn{1}{l|}{0.76}    & 0.74  & \multicolumn{1}{l|}{0.82}    & 0.81 \\ \hline
Salience Pred.+IndicBert & \multicolumn{1}{l|}{0.76}          &  0.74       & \multicolumn{1}{l|}{0.80}          & 0.78       \\ \hline
Multi-Task                      & \multicolumn{1}{l|}{\textbf{0.78}}    & \textbf{0.77}  &\multicolumn{1}{l|}{\textit{0.80}}    & \textit{0.78}  \\ \hline
\end{tabular}
\caption{Model results. For TF-IDF and TextRank models we take the sum of the token embeddings.}
\label{tab:model}
\end{table}

\noindent\textbf{Implementation Details:} All models are trained using GeForce RTX 2080Ti GPUs. Models are tuned for hyper-parameters using the validation set (details in Appendix \ref{app:hyperparameter}). 

\subsection{Results}
The results are shown in Table \ref{tab:model}. As can be observed, in general, the performance of models is lower in the case of district-wise settings. This is possibly due to the lexical variation (section \ref{sec:hldc}) across districts, which makes it difficult for the model to generalize. Moreover, this lexical variation corresponds to the usage of words corresponding to dialects of Hindi. 
Another thing to note from the results is that, in general, summarization based models perform better than Doc2Vec and transformer-based models, highlighting the importance of the summarization step in the bail prediction task. The proposed end-to-end multi-task model outperforms all the baselines in the district-wise setting with 78.53\% accuracy. The auxiliary task of sentence salience classification helps learn robust features during training and adds a regularization effect on the main task of bail prediction, leading to improved performance than the two-step baselines. However, in the case of an all-district split, the MTL model fails to beat simpler baselines like TF-IDF+IndicBERT. We hypothesize that this is due to the fact that the sentence salience training data may not be entirely correct since it is based on the cosine similarity heuristic, which may induce some noise for the auxiliary task. Additionally, there is lexical diversity present across documents from different districts. Since documents of all districts are combined in this setting, this may introduce diverse sentences, which are harder to encode for the salience classifier, while TF-IDF is able to look at the distribution of words across all documents and districts to extract salient sentences. 



\subsection{Error Analysis}
\begin{figure}[!h]
\includegraphics[width=0.85\columnwidth]{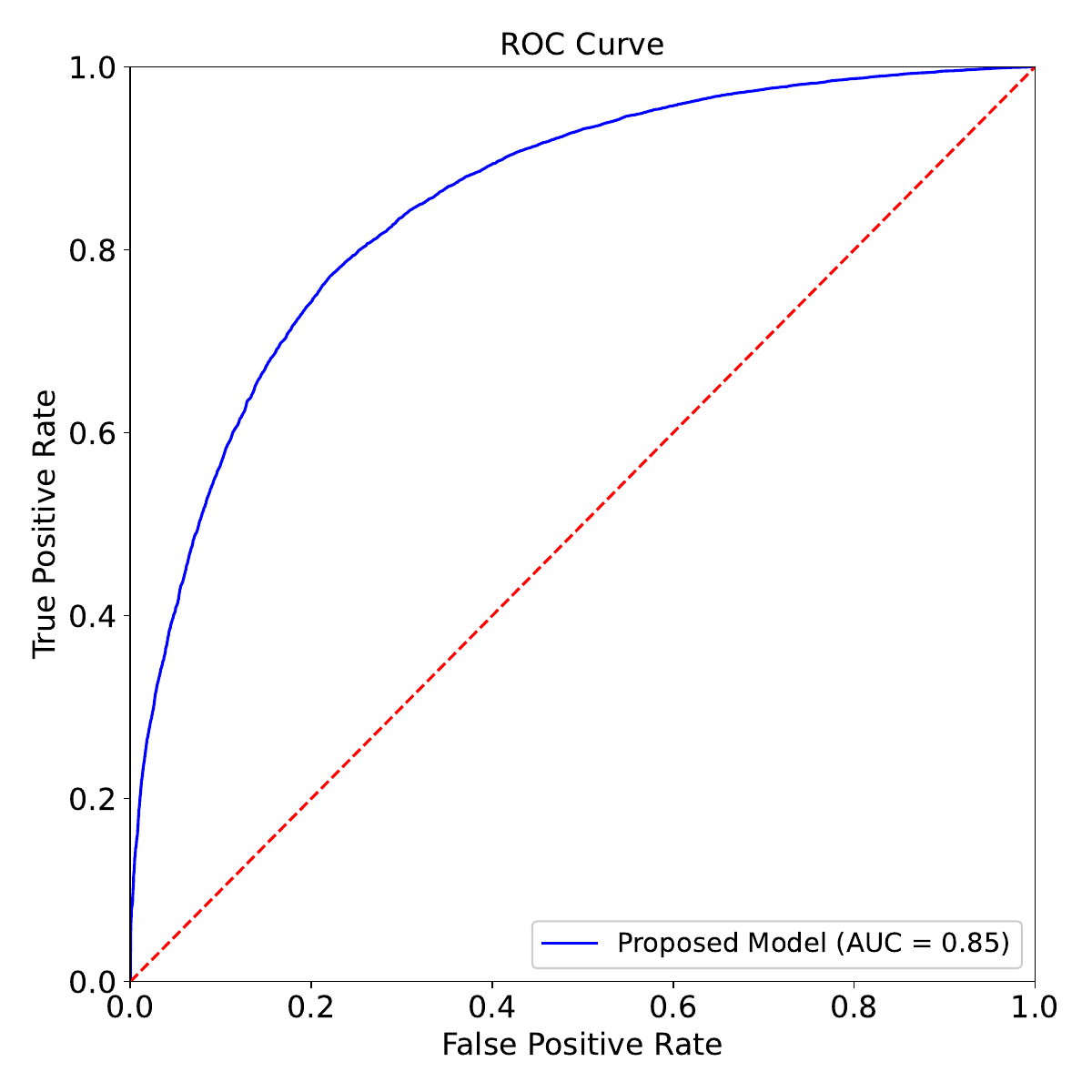}
\centering
\caption{ROC curve for the proposed model. The total AUC (Area under curve) is 0.85.} 
\label{fig:bail-roc}
\end{figure}

We did further analysis of the model outputs to understand failure points and figure out improvements to the bail prediction system. After examining the miss-classified examples, we observed the following. First, the lack of standardization can manifest in unique ways. In one of the documents, we observed that all the facts and arguments seemed to point to the decision of bail granted. Our model also gauged this correctly and predicted bail granted. However, the actual result of the document showed that even though initially bail was granted because the accused failed to show up on multiple occasions, the judge overturned the decision and the final verdict was bail denied. In some instances, we also observed that even if the facts of the cases are similar the judgements can differ. We observed two cases about the illegal possession of drugs that differed only a bit in the quantity seized but had different decisions. The model is trained only on the documents and has no access to legal knowledge, hence is not able to capture such legal nuances. 
We also performed quantitative analysis on the model output to better understand the performance. Our model outputs a probabilistic score in the range $\{0, 1\}$. A score closer to 0 indicates our model is confident that bail would be denied, while a score closer to 1 means bail granted. In Figure \ref{fig:bail-roc} we plot the ROC curve to showcase the capability of the model at different classification thresholds. ROC plots True Positive and False Positive rates at different thresholds. The area under the ROC curve (AUC) is a measure of aggregated classification performance. Our proposed model has an AUC score of 0.85, indicating a high-classification accuracy for a challenging problem.  

\begin{figure}[!h]
\includegraphics[scale=0.4]{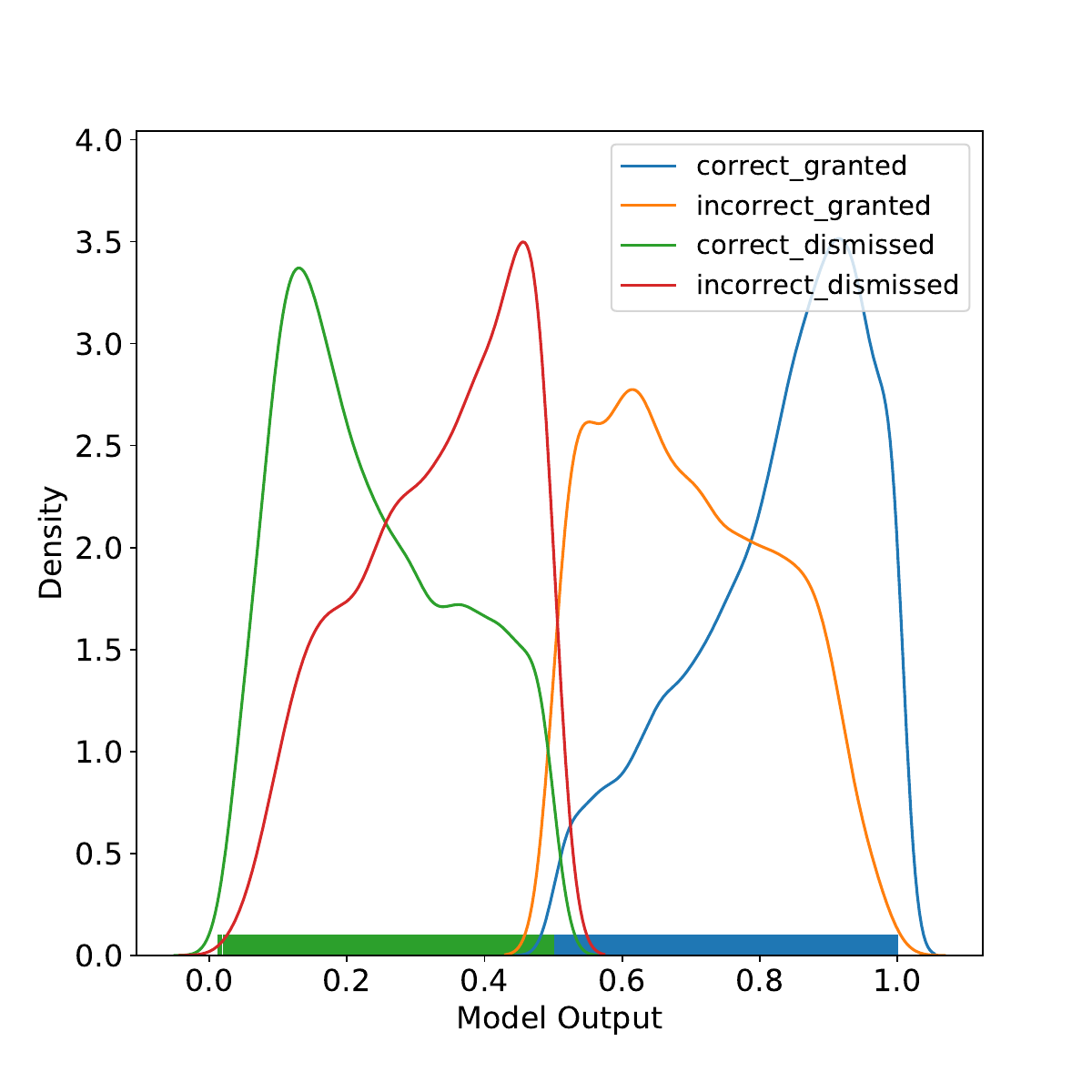}
\centering
\caption{Kernel Density Estimate (KDE) plots of our proposed bail prediction model. The majority of errors (incorrectly dismissed / granted) are borderline cases with model output score around 0.5.} 
\label{fig:bail-density}
\end{figure}

We also plot (Figure \ref{fig:bail-density}) the density functions corresponding to True Positive (Bail correctly granted), True Negative (Bail correctly dismissed), False Positive (Bail incorrectly granted) and False Negatives (Bail incorrectly dismissed). We observe the correct bail granted predictions are shifted towards 1, and the correct bail denied predictions are shifted towards 0. Additionally, the incorrect samples are concentrated near the middle ($\approx$ 0.5), which shows that our model was able to identify these as borderline cases. 

\section{Future Work and Conclusion} \label{conclusion}

In this paper, we introduced a large corpus of legal documents for the under-resourced language Hindi: Hindi Legal Documents Corpus (HLDC). We semi-structure the documents to make them amenable for further use in downstream applications. As a use-case for HLDC, we introduce the task of Bail Prediction. We experimented with several models and proposed a multi-task learning based model that predicts salient sentences as an auxiliary task and bail prediction as the main task. Results show scope for improvement that we plan to explore in future. We also plan to expand HLDC by covering other Indian Hindi speaking states. Furthermore, as a future direction, we plan to collect legal documents in other Indian languages. India has 22 official languages, but for the majority of languages, there are no legal corpora. Another interesting future direction that we would like to explore is the development of deep models infused with legal knowledge so that model is able to capture legal nuances. We plan to use the HLDC corpus for other legal tasks such as summarization and prior case retrieval. 

    

\section{Acknowledgements}\label{sec:ack}

This paper is dedicated to T.H. Arjun, who contributed towards making this research possible, you will be remembered! We would like to thank  Prof. Angshuman Hazarika and Prof. Shouvik Kumar Guha for their valuable suggestions and for guiding us regarding the technical aspects of the Indian legal system. The author Ashutosh Modi would like to acknowledge the support of Google Research India via the Faculty Research Award Grant 2021. This project was partially supported by iHub at IIIT Hyderabad, project O2-001.

\bibliography{references}
\bibliographystyle{acl_natbib}

\clearpage
\newpage
\appendix
\section*{Appendix}

\section{Data Statistics}
\begin{table}[h]
    \centering
    \small
    \begin{tabular}{|c|c|}
        \hline
        District & Number of Bail Applications \\
        \hline
        Muzaffarnagar & 17234 \\
        Moradabad & 16219 \\
        Budaun & 14533 \\
        Sitapur & 14478 \\
        Saharanpur & 10838 \\
        \hline
    \end{tabular}
    \caption{Top 5 districts with most number of bail applications in UP.}
    \label{table:bail_cases_per_district}
\end{table}

\section{Data Cleaning and Filtering}\label{app:data-cleaning}
1,221,950 documents were scraped from Ecourts website and 309,382 documents were removed in the cleaning and filtering process. Following rules were used to remove documents.
\begin{itemize}
    \item Removed blank documents (whose length is less than 32 bytes)
    \item Removed duplicate documents
    \item Removed too long and too short documents (>8096 bytes or <2048 bytes). 
    \item Removed document where majority text was in English language.
\end{itemize}
This resulted in 912,568 filtered case documents that  constitute the Hindi Legal Document Corpus. 

\section{NER Removal} \label{ner-removal}
For removing names and locations, lookup was done in lists containing NER. Libraries like HindiNLP\footnote{\url{https://github.com/avinsit123/HindiNLP}} (which  uses SequenceTagger from flair library\footnote{\url{https://github.com/flairNLP/flair}} which is based on an RNN model) were run on a subset of the data to find additional NER that were added to the lists. Since the Sequence-Tagger model is quite slow in processing documents, directly tagging  large HLDC is not efficient. 
If a word was found in one of these lists then it was replaced with a \textless{\dn nAm}\textgreater{} (<name>) tag. Phone numbers were replaced with \textless{\dn \327won{\rs -\re}n\2br}\textgreater{} (<phone-number>) tag using the following regex 
\begin{lstlisting}
((\+*)((0[ -]*)*|((91 )*))((\d{12})
+|(\d{10})+))|\d{5}([- ]*)\d{6}
\end{lstlisting}
Phone numbers written in Hindi were also considered by using the same regex as above with English digits replaced with Hindi ones. 

\section{Document Segmentation} \label{app:bail-document-segmentation} 
Out of 912,568 documents in HLDC, 340,280 were bail documents, these were further processed to obtain the Bail Document corpus. Bail documents were structured into different sections. We extracted these sections from the bail documents. Details are mentioned below.
An example of document with different sections is shown in Table \ref{tab:long}. 

\subsection{Header}\label{app:bail-header}
Header refers to the meta data related to the case, for example, {\dn DArA} (IPC (Indian Penal Code) sections), {\dn TAnA} (police station), case number, date of hearing, accused name, etc. Header is present at the top of the document. Header mostly ended with {\dn DArA} (IPC) or {\dn TAnA} (police station) details. Hence, in order to cut the document to get header, we first find the indices of {\dn DArA} (IPC) and {\dn TAnA} (police station), and from these indices we find the finishing word of the header. We then segment the document at the finishing word. We also include the  first line of upcoming paragraph in header as it also didn't contain case arguments but contained data like if this is the first bail application or not.

\subsection{Case Result}\label{app:bail-result}
Case Result refers to the end of the document where judge writes their decision. Judge either accepts the bail application or rejects it. If the judge had accepted the bail document then this section mostly also contains bail amount and bail terms for accused. \\
We observed that result section mostly began along the following line, {\dn mAml\? k\? sm-t tLyo{\qva} ko d\?Kkr} (looking at all facts of the case), the keyword {\dn tLyo{\qva}} (facts) was very common around the start of the result section. Hence, we iterated over the indices of keyword {\dn tLyo{\qva}} (facts) in reverse order and checked if the division at that index is correct. To check if the division is correct we look for bail result in lower half of the division, if the bail result is present, we classify that division as correct else we move to next index of {\dn tLyo{\qva}} (facts).

\subsection{Body}\label{app:bail-body}
The remaining portion of the document after removing header and result section was called body. Body section was further divided, as described below.

\subsubsection{Judge's summary}\label{app:bail-judge}
Most of the bail documents have a concluding paragraph where the judge summarizes their viewpoints of the case. To extract this, we first constructed certain regex which often precedes judge's summary, defendant's and prosecutor's arguments (described in Table ~\ref{tab:judges-summary}). Since the document might have intermingling of different arguments and opinions, we opted for sentence level annotation of these labels using the regex pattern. The sentences not matching any criteria are given a tag of None. Next we try to replace the None by extending the tags of the sentences to paragraph level as long as no other tag is encountered.
As the judge's opinion mostly occurs at the end, we start iterating from end and start marking the None as judge's opinion. If a label which is neither None nor judge's opinion is encountered, the document is discarded as we cannot extract the judge's opinion from the document using the process defined. If the judge's opinion label is found in reverse iteration, then we claim that judge's opinion can be extracted. Finally, all sentences labelled as judge's opinion either during reverse iteration or during paragraph level extension are extracted out as judge's summary and rest of the sentences form facts and opinions for further modelling.
Using the above process, following are some cases where the judge's opinion cannot be extracted:
\begin{enumerate}[parsep=0pt]
    \itemsep0em
    \item Certain characters were mis-identified in the OCR pipeline and hence do not match the regex. 
    \item The segmentation of document into header, body and result caused a significant portion of the body and thus judge's opinion to move to result section.
    \item The document was written from judge's perspective and hence judge's summary also contains the prosecutor's and defendant's arguments.
    \item The regex didn't have 100\% coverage.
\end{enumerate}

\begin{table}[h]
\small
    \centering
    \begin{tabular}{|p{0.17\columnwidth}|p{0.32\columnwidth}|p{0.3\columnwidth}|}
        \hline
        Field & Hindi phrases & English Translations\\ 
        \hline
        Judge's Summary & {\dn uBy p\322w kF bhs \7{s}nn\?{\rs ,\re} p/AvlF k\? avlokn{\rs ,\re} k\?s XAyrF m\?{\qva} uplND sA\323wy k\? a\7{n}sAr{\rs ,\re} mAml\? k\? tLyo{\qva} v pErE-TEtyo{\qva} m\?{\qva} \8{p}rF trh s\? -p\3A3w h\4{\rs ,\re} \3FEwTm \8{s}cnA ErpoV\0{\rs ,\re} \7{p}Els \3FEwp/} \ldots {\dn prFEfln EkyA} & Hearing the arguments of the parties, perusal of the records, as per the evidence available in the case diary, fully clear from the facts and circumstances of the case, First Information Report, Police Forms \ldots perused\\ \hline
        Prosecutor & {\dn jmAnt kA EvroD krt\? \7{h}y\? aEByojn kF aor s\? tk\0 EdyA gyA h\4{\rs ,\re} jmAnt \3FEwAT\0nAp/ k\? Ev!\388w aApE\381w} & Opposing the bail, it has been argued on behalf of the prosecution, the objection against the bail application\\ \hline
        Defendant & {\dn aEB\7{y}\3C4w k\? Ev\392wAn aEDv\3C4wA kA tk\0 h\4{\rs ,\re} m\?{\qva} \8{J}WA ev\2 r\2Ejfn P\2sAyA gyA} & The learned counsel for the accused has argued, has been falsely and enmity implicated in this case\\ \hline
    \end{tabular}
    \caption{Phrases used to construct regular expression for extracting judge's opinion. The list is just an indicative of the various phrases and variants used; the entire list can be found in code}
    \label{tab:judges-summary}
\end{table}

\subsubsection{Facts and Arguments}\label{app:bail-facts}
This section comprised of facts related to case, arguments from defendant and prosecutor. Mostly, this corresponds to the portion of the body after removing judge's summary. 

\section{Extracting Bail Decision from Result}\label{app:bail-decision}
\label{section:appendix_bail_decision_result}
To extract the bail decision we searched for keywords in result section. Keywords like {\dn KAErj} (dismissed) and {\dn Enr-t} (invalidated) identified rejection of bail application and words like {\dn -vFkAr} (accepted) identified acceptance of bail application. Table \ref{table:bail_decision_tokens} lists all the tokens used for extraction.

\begin{table}[h]
\small
        \centering
        \begin{tabular}{|m{0.2\columnwidth}|m{0.7\columnwidth}|}
            \hline  
            Field & Tokens\\
            \hline
            Bail granted tokens
            & \begin{minipage}{.2\textwidth}
            \includegraphics[scale=0.15]{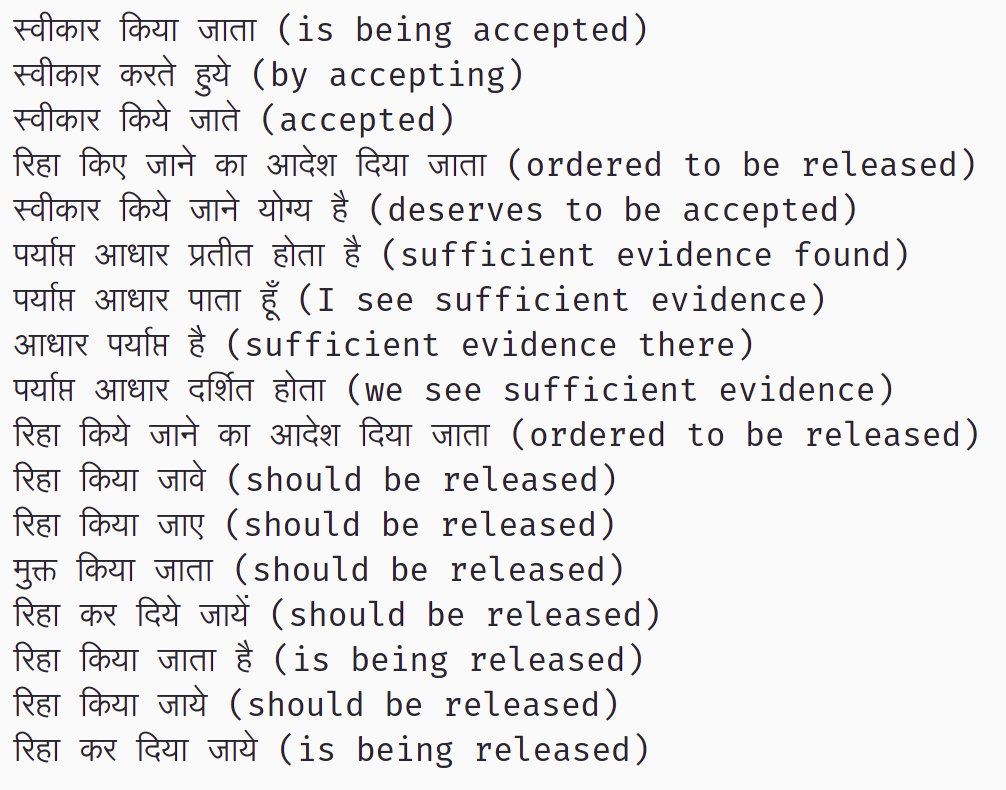}
            \end{minipage}\\
            \hline
            Bail denied tokens
            & \begin{minipage}{.2\textwidth}
            \includegraphics[scale=0.14]{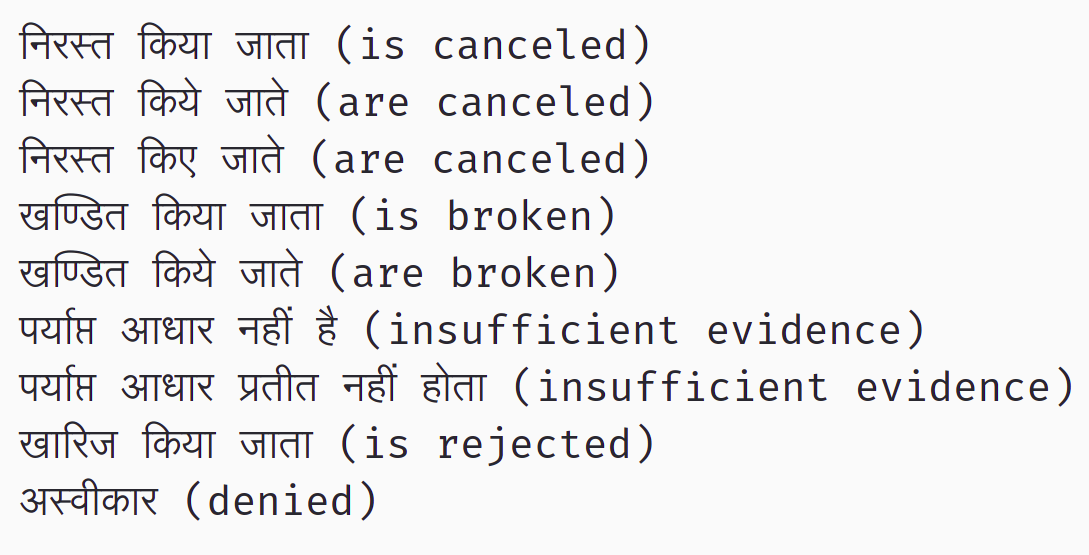}
            \end{minipage} \\
            \hline
        \end{tabular}
        \caption{Bail decision tokens}
        \label{table:bail_decision_tokens}
\end{table}


\section{Extracting Bail Amount from Result}\label{app:bail-amount}
\label{section:appendix_bail_amount_extraction}
In case of granted bail decision, the judge specifies bail amount. We saw that the bail amount mostly comprises of personal bond money and surety money. There can be multiple personal bonds and sureties. The bail amount we extracted refers to the sum of all the personal bond money.
Bail amount was present in two forms in result section, numerical and Hindi-text. Numerical bail amount was extracted by regex matching and text bail amount was extracted by creating a mapping for it. Table \ref{table:bail_amount_mapping} shows few examples of bail mapping.

\section{HLDC Pipeline Analysis}\label{app:pipeline}
We used a validation set (0.1\% of data) to evaluate our regex based approaches, the results are in Table \ref{table:pipeline_results}. Note that metrics used for evaluation are quite strict and hence the results are much lower for Judge's summary part. The segmentation and Judge's opinion were strictly evaluated and even a single sentence in the wrong segment reduces the accuracy. We also see that the main binary label of outcome detection (bail granted or denied) had an almost perfect accuracy of 99.4\%. Nevertheless,  in future we plan to improve our pipeline further by training machine learning models.  


\begin{table}[h]
    \centering
    \small
    \begin{tabular}{|m{13em}|m{4em}|}
        \hline
        Process & Accuracy \\
        \hline
        Header, Body and Case Result Segmentation & 89.7\% \\
        \hline
        Judge's Opinion and Facts extraction & 85.7\% \\
        \hline
        Bail Decision Extraction & 99.4\% \\
        \hline
    \end{tabular}
    \caption{Evaluation results of bail document division and bail decision extraction pipeline.}
    \label{table:pipeline_results}
\end{table}

\begin{table}[h]
        \centering
        \small
        \begin{tabular}{|p{30mm}|p{20mm}|}
            \hline  
            Text Amount & In Value Form\\
            \hline
            \begin{minipage}{.2\textwidth}
            \includegraphics[width=32mm, height=25mm]{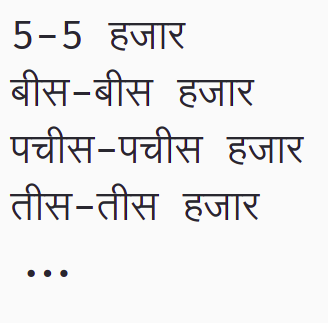}
            \end{minipage}
            & \begin{minipage}{.2\textwidth}
            \includegraphics[width=18mm, height=20mm]{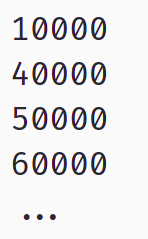}
            \end{minipage}\\
            \hline
            
        \end{tabular}
        \caption{Text bail amount mapping example}
        \label{table:bail_amount_mapping}
\end{table}

\section{Model Details}
\subsection{Evaluation Metrics} \label{eval-metrics}
To evaluate the performance of all the models, we use Accuracy, and F1-score, which are considered standard evaluation metrics while performing classification experiments. These are mathematically described as the follows: 
\begin{align*}
Accuracy =& \frac{TP + TN}{TP + TN + FP + FN} \\ \\    
F_1\ Score =& \frac{2 * Precision * Recall}{ Precision + Recall}     
\end{align*}


    
where TP, FP, TN, and FN denote True Positives, False Positives, True Negatives, and False Negatives, respectively. The mathematical formulation for $Precision$ and $Recall$ is given as follows:
$$
    Precision = \frac{TP}{TP + FP} 
$$
$$
    Recall = \frac{TP}{TP + FN} 
$$

\subsection{Hyperparamter Tuning} \label{app:hyperparameter}

\begin{table}[]
\small
\centering
\begin{tabular}{|p{0.2\columnwidth}|p{0.30\columnwidth}p{0.30\columnwidth}|}
\hline
Model                      & \multicolumn{2}{p{0.66\columnwidth}|}{Hyper-Parameters (L=Learning Rate), (E=Epochs), (D=Embedding Dimension(Default 200)), (W= Weight Decay), (E=Epochs(Default 15))} \\ \cline{2-3} 
                           & \multicolumn{1}{l|}{District-wise Split}    & All Districts Split   \\ \hline
Doc2Vec + SVM                & \multicolumn{1}{p{0.30\columnwidth}|}{E=100}           & E=100         \\ \hline
Doc2Vec + XGBoost            & \multicolumn{1}{p{0.30\columnwidth}|}{E=100, D=300}                       &   E=100, D=300                   \\ \hline
IndicBert - (First 512 Tokens)     & \multicolumn{1}{p{0.30\columnwidth}|}{L=\num{3.69e-6}, W=\num{2.6e-2}}   & L=\num{1.58e-06}, W=\num{4.8e-2}                     \\ \hline
IndicBert - (Last 512 Tokens)       & \multicolumn{1}{p{0.30\columnwidth}|}{L=\num{5.60e-5}, W=\num{1.0e-2}}   & L=\num{2.18e-05}, W=\num{4.3e-2}                     \\ \hline
TF-IDF + IndicBert         & \multicolumn{1}{p{0.30\columnwidth}|}{L=\num{1.11e-5}, W=\num{1.9e-2}}   & L=\num{9.84e-06}, W=\num{4.9e-2}                     \\ \hline
TextRank + IndicBert       & \multicolumn{1}{p{0.30\columnwidth}|}{L=\num{3.17e-06}, W=\num{3.1e-2}}  & L=\num{3.99e-06}, W=\num{1.5e-2}                       \\ \hline
Salience Pred. + IndicBert & \multicolumn{1}{p{0.30\columnwidth}|}{L=\num{1e-5}, W=\num{3.2e-2}}                       &  L=\num{4.2e-06}, W=\num{1.7e-2}                     \\ \hline
Multi-Task                 & \multicolumn{1}{p{0.30\columnwidth}|}{E=30, L=\num{5e-5}}          &  E=30, L=\num{1e-5}                   \\ \hline
\end{tabular}
\caption{Listing of Hyper-Parameters for Training of Models}
\label{tab:paramter_tuning}
\end{table}
We used Optuna \footnote{\url{https://github.com/optuna/optuna}} for hyperparameter optimisation. Optuna allows us to easily define search spaces, select optimisation algorithms and scale with easy parallelization. We run parameter tuning on 10\% of the data to identify the best parameters before retraining the model with the best parameters on the entire dataset. The best parameters are listed in Table \ref{tab:paramter_tuning}.
\onecolumn
\begin{center}
\section{Sample Segmented Document}
\label{app:sample-example}
\begin{longtable}{|p{0.19\textwidth}|p{0.33\textwidth}|p{0.37\textwidth}|}
\hline \multicolumn{1}{|c|}{\textbf{Field}} & \multicolumn{1}{c|}{\textbf{Example}} & \multicolumn{1}{c|}{\textbf{Translation}} \\ \hline 
\endfirsthead
\multicolumn{3}{c}%
{{\bfseries \tablename\ \thetable{} -- continued from previous page}} \\
\hline \multicolumn{1}{|c|}{\textbf{Field}} & \multicolumn{1}{c|}{\textbf{Example}} & \multicolumn{1}{c|}{\textbf{Translation}} \\ \hline 
\endhead

\hline \multicolumn{3}{|r|}{{Continued on next page}} \\ \hline
\endfoot
\endlastfoot

\textbf{Header}: This chunk of the document contains meta information related to the case like court hearing date, IPC sections attached, police station of complain, etc.
& \begin{minipage}{.2\textwidth}
\includegraphics[width=55mm, height=85mm]{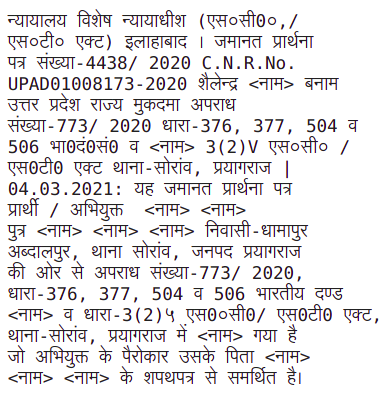}
\end{minipage}
& \begin{minipage}{.2\textwidth}
\includegraphics[width=60mm, height=85mm]{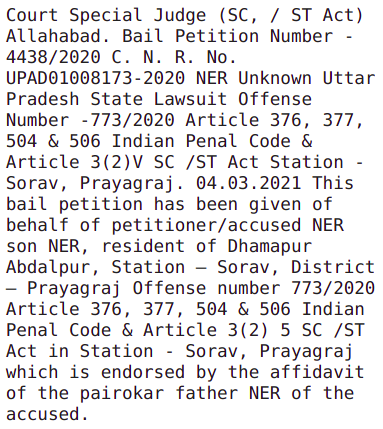}
\end{minipage} \\
\hline
\textbf{Facts and Arguments}: This chunk of the document contains case facts related to the case and arguments from defendant and prosecutor.
& \begin{minipage}{.2\textwidth}
\includegraphics[width=55mm, height=110mm]{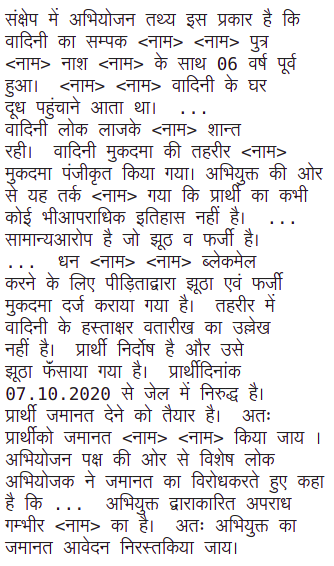}
\end{minipage}
& \begin{minipage}{.2\textwidth}
\includegraphics[width=60mm, height=125mm]{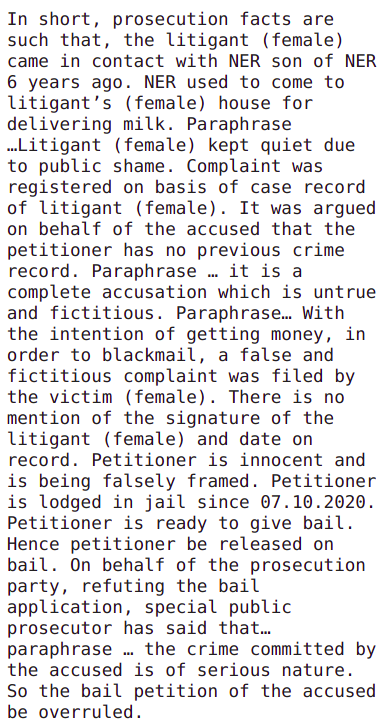}
\end{minipage} \\
\hline
\textbf{Judge's Opinion}: This refers to the few lines present in the middle portion of the document where judge writes their opinion of the case. & 
\begin{minipage}{.2\textwidth}
\includegraphics[width=55mm, height=40mm]{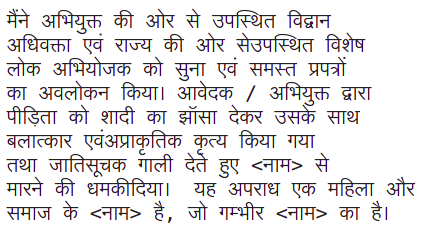}
\end{minipage} &
\begin{minipage}{.2\textwidth}
\includegraphics[width=60mm, height=50mm]{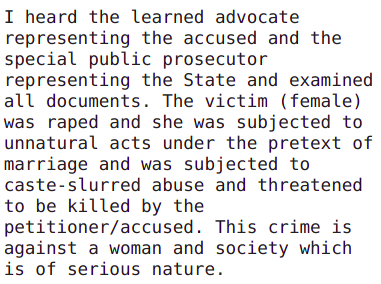}
\end{minipage}\\
\hline
\textbf{Result}: This chunk of the document contains decision made by judge on the case.
& \begin{minipage}{.2\textwidth}
\includegraphics[width=55mm, height=50mm]{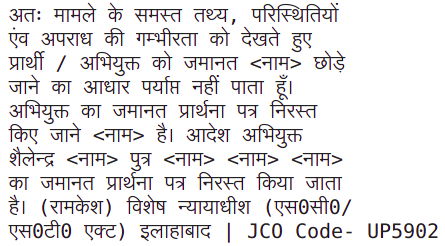}
\end{minipage}
& \begin{minipage}{.2\textwidth}
\includegraphics[width=60mm, height=50mm]{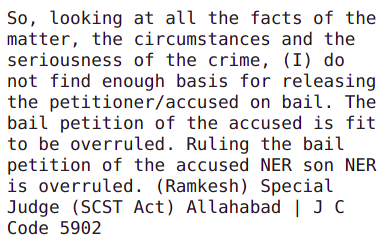}
\end{minipage} \\
\hline
\caption{A sample segmented document} \label{tab:long}\\
\end{longtable}
\end{center}
\clearpage
\twocolumn

\end{document}